\documentclass{article}

\usepackage{arxiv}

\usepackage[utf8]{inputenc} % allow utf-8 input
\usepackage[T1]{fontenc}    % use 8-bit T1 fonts
\usepackage{hyperref}       % hyperlinks
\usepackage{url}            % simple URL typesetting
\usepackage{booktabs}       % professional-quality tables
\usepackage{amsfonts}       % blackboard math symbols
\usepackage{amsmath}
\usepackage{nicefrac}       % compact symbols for 1/2, etc.
\usepackage{microtype}      % microtypography
\usepackage{lipsum}
\usepackage{graphicx}
\graphicspath{ {./images/} }

\usepackage{natbib}

\title{Grid Cell Path Integration For Movement-Based Visual Object Recognition}

\author{
 Niels Leadholm \\
  Numenta and The University of Oxford\\
  \texttt{niels.leadholm@seh.ox.ac.uk} \\
   \And
 Marcus Lewis \\
  Numenta \\
  \texttt{mlewis@numenta.com} \\
  \And
 Subutai Ahmad \\
  Numenta \\
  \texttt{sahmad@numenta.com} \\
}

\begin{document}
\maketitle
\begin{abstract}
Grid cells enable the brain to model the physical space of the world and navigate effectively via path integration, updating self-position using information from self-movement. Recent proposals suggest that the brain might use similar mechanisms to understand the structure of objects in diverse sensory modalities, including vision. In machine vision, object recognition given a sequence of sensory samples of an image, such as saccades, is a challenging problem when the sequence does not follow a consistent, fixed pattern - yet this is something humans do naturally and effortlessly. We explore how grid cell-based path integration in a cortical network can support reliable recognition of objects given an arbitrary sequence of inputs. Our network (GridCellNet) uses grid cell computations to integrate visual information and make predictions based on movements. We use local Hebbian plasticity rules to learn rapidly from a handful of examples (few-shot learning), and consider the task of recognizing MNIST digits given only a sequence of image feature patches. We compare GridCellNet to k-Nearest Neighbour (k-NN) classifiers as well as recurrent neural networks (RNNs), both of which lack explicit mechanisms for handling arbitrary sequences of input samples. We show that GridCellNet can reliably perform classification, generalizing to both unseen examples and completely novel sequence trajectories. We further show that inference is often successful after sampling a fraction of the input space, enabling the predictive GridCellNet to reconstruct the rest of the image given just a few movements. We propose that dynamically moving agents with active sensors can use grid cell representations not only for navigation, but also for efficient recognition and feature prediction of seen objects.
\end{abstract}

% keywords can be removed

\section{Introduction}

When exploring a visual scene, primates sample the world in a serial sequence by performing saccades \citep{Yarbus1967EyeObjects}. For the purpose of recognising objects, it is non-trivial that this sampling can follow an arbitrary sequence order. For example, one might selectively attend to the most salient parts of a face rather than performing a raster scan across the image. While many previous efforts to model primate object recognition have focused on massively parallel processing of a single input, the challenge of dealing with the necessarily sequential nature of sensory input has received less attention \citep{Bicanski2019ACells}. Recurrent-neural networks can perform complex tasks with sequential inputs, and might seem like a natural candidate for such a challenge, yet they struggle to learn when provided with sequences that do not follow a fixed order during training and inference. In the natural world there are additional challenges that can present themselves. Often only a handful of object examples are available, and learning should be rapid (i.e. requiring limited training on the few examples given). These are all constraints that humans are able to handle effortlessly. Understanding how learning and inference under these conditions might be achieved has two appealing aspects. As well as potentially uncovering the basis for human performance in this domain, the flexibility to operate under such a regime could also enable artificial agents to explore the world in a more principled and adaptive manner.

While recurrent neural networks do not have explicit mechanisms for dealing with this challenge, grid cells might provide a neurally plausible solution employed by the brain. Together with place cells in the hippocampus \citep{OKeefe1978HippocampalFire}, grid cells in the entorhinal cortex enable the brain to model space during navigation. In particular, grid cells fire in repeating patterns as space is traversed \citep{Hafting2005MicrostructureCortex}. Using multiple grid cells of different scale and orientation, the location of an animal can be uniquely encoded \citep{Fiete2008WhatLocation}. Importantly, this location representation can be updated to support path integration - that is, given information about self-movement, an agent can determine its new location by reading out from grid cell activity \citep{Hafting2005MicrostructureCortex, Moser2008PlaceSystem}. The role of such cells in spatial navigation is widely established, but recent experimental evidence has also uncovered the presence of grid cell-like activity in visual space \citep{Killian2012ACortex, Nau2018HexadirectionalCortex, Julian2018HumanGrid}. Theoreticians have argued that grid cell-like computations might be used to build object representations in diverse sensory modalities \citep{Hawkins2019ANeocortex}, including vision \citep{Bicanski2019ACells}. This is an intriguing solution to our opening problem, but the demonstration of object recognition with such computations has so far been limited to either synthetic objects \citep{Lewis2019LocationsCells}, or visual tasks requiring the recall of a memorized example \citep{Bicanski2019ACells}, rather than generalization to unseen examples of an object class.

Neurally-motivated systems that can solve rapid object learning and recognition given saccade-like visual inputs are therefore lacking. We set out to address this by implementing a biologically plausible network, called GridCellNet, based on cortical columns and grid cell-like computations. The system uses rapid Hebbian-style learning to associate sensed features and their spatial location in the reference frame of an object, while dendritic segments enable the system to encode predictive states. Locations are encoded by activity in grid cell modules that are updated with self-movement. In addition to addressing the challenge of arbitrary sequence inputs, this system also includes the desirable properties of rapid learning (functioning with both few training examples and few weight updates), and predictive capabilities, enabling completion of an image given partial inputs. We evaluate the performance of GridCellNet in these task settings, and compare it to typical machine-learning approaches. In accordance with human capabilities, our system outperforms these other approaches in the challenging setting we explore. To summarise, our primary contributions are to:

\begin{itemize}
    \item Implement a biologically-motivated neural architecture that uses arbitrary sequences of local visual features across space to learn objects and recognize them. This ability is dependent on grid cell-like computations.
    
    \item Demonstrate the ability of our network to successfully generalize to unseen objects in the challenging setting of arbitrary sequence inputs under few-shot learning.

    \item Compare the proposed system to more traditional machine learning systems, such as recurrent neural networks, demonstrating its superiority in settings where humans outperform current artificial systems.
\end{itemize}

\section{Related Literature}

\begin{figure}[ht]
\begin{center}
\includegraphics[width=0.9\textwidth]{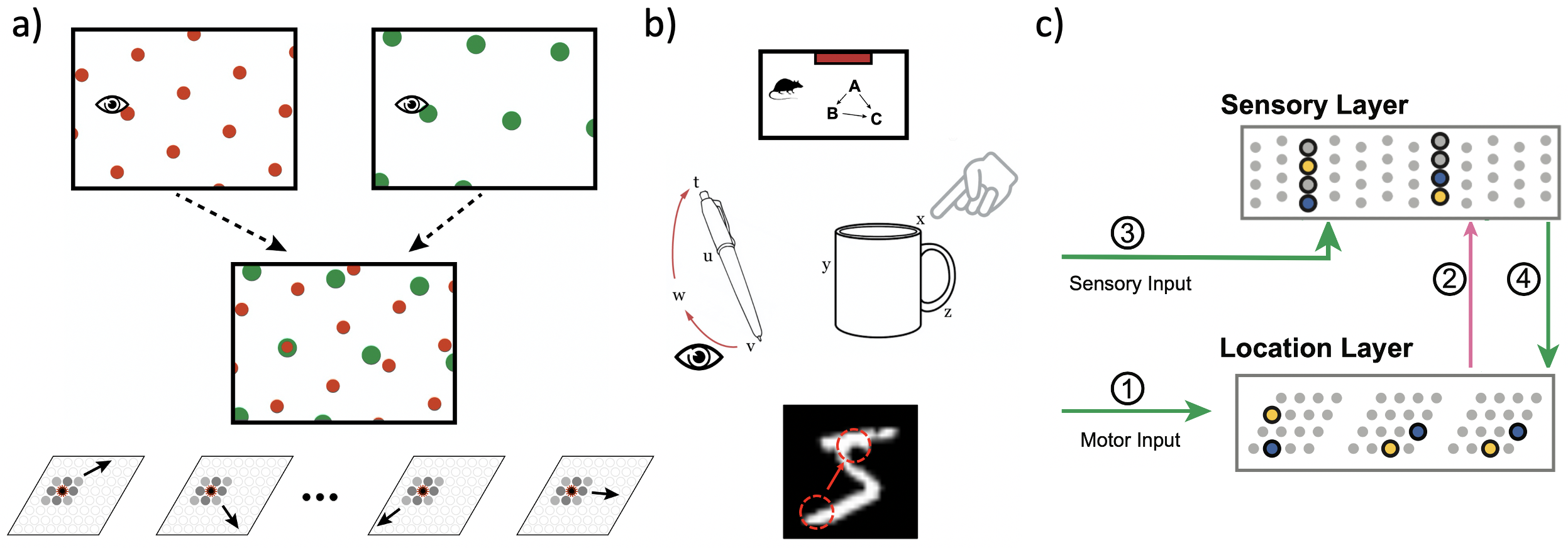}
\end{center}
\caption{\label{fig:general_intuition}\textit{Using grid cell representations for object recognition.} a) The combination of multiple grid cells of different scale and orientation (red and green) can uniquely encode the location of a sensor (e.g. retinal patch). Here we use multiple grid cell modules with sparse activity (bottom, each indicated by a rhombus) to encode and update the sensor’s location with self-movement information. b) We hypothesize that this process can be used for object recognition with active sensors, and use sequences through a 5x5 grid of feature patches extracted from MNIST images to test this. c) GridCellNet takes in motor input when the sensor moves (1) and updates its location representations. The current location representation is used to predict incoming sensory information (2), before this is received (3). Correctly predicted sensory information is then used to update the location representation (4). Locations are initially ambiguous, represented using a union of locations, and disambiguated over time via sensory input. Yellow and blue dots in the location layer indicate two different objects which are compatible with the current sequence. Classification is successful once the representation is unambiguous, i.e. that it is a subset of representations of that class. The two-layer network is based on cells with sparse binary activity, dendritic segments, and Hebbian-like learning, following \citet{Lewis2019LocationsCells}, and with figures reproduced with permission from the authors.}
\end{figure}

\textbf{Grid Cells} During spatial navigation in an animal such as a rat, grid cells are notable for firing at regular intervals as space is traversed. These points of activity correspond to a triangular lattice with a particular phase, orientation and scale \citep{Hafting2005MicrostructureCortex} (Figure \ref{fig:general_intuition} top). Grid cells with the same orientation and scale, but different phases, form what are known as grid cell `modules' \citep{Stensola2012TheDiscretized}. As a rodent moves, any individual grid cell's activity is ambiguous as a means of encoding the animal's position. The joint activity of multiple grid cell modules however can uniquely encodes a position; importantly, this encoding scheme has a large representational capacity \citep{Fiete2008WhatLocation} (Figure \ref{fig:general_intuition}a middle). Information about self-movement is used to update the current location representation by each grid cell's firing corresponding to the positional change (Figure \ref{fig:general_intuition}b bottom). This process, known as path-integration, means that after returning to the same position, the same grid cells will be active regardless of the path taken \citep{Hafting2005MicrostructureCortex, Moser2008PlaceSystem}. Note that our work does not deal with how grid cells might actually implement path integration - rather we explore the significance of path integration in a neural population for developing useful object representations. 

The combined properties of a large capacity for unique spatial representations and path integration enable grid cells to act as a powerful substrate for encoding spatial information. In \citet{Hawkins2019ANeocortex}, it was suggested that similar cells might exist outside of the entorhinal cortex in cortical columns throughout the brain, and could thereby support spatial encoding in sensory modalities such as touch and vision. Recent experimental work has supported this possibility \citep{Long2021ACortex, Long2021AFormation}. This work aims to demonstrate the utility of such grid cell representations in visual tasks. For a more in depth discussion of grid cell computations as explored here, we direct readers to \citet{Lewis2019LocationsCells}. 

\textbf{Cortical Models} Our work builds on previous models of the cortical architecture of the mammalian brain. This prior work demonstrated that networks with a columnar architecture, where different layers correspond to sensory and location-based representations, can learn inputs such as objects composed of synthetic features. Neurons in these layers receive external sensory and self-movement information, while they share connections that enable learned associations between features and locations, as well as predictions during inference \citep{Hawkins2017}. Neural activity (including input features) are represented in the distributed activity of sparse binary vectors - a form of encoding where the dimensionality used is relatively high, but where only a small subset of given nodes are ever active at a time (taking values of 0 or 1). Such encoding has numerous appealing properties, including tolerance to noise \citep{Ahmad2019HowRepresentations}, a large representational capacity, and the ability to encode notions of similarity between objects \citep{Hawkins2016WhyNeocortex}. 

Early models in this framework did not explicitly discuss how the brain might implement the encoding of location information. It was recently suggested that this might be provided by neurons with grid cell properties distributed throughout the cortical columns of the brain. In various sensory modalities, grid cells could then be used to encode feature locations in an object's own reference frame. The idea that each of the columns throughout the brain would be learning object representations in a massively parallel process was dubbed the Thousand Brains Theory. This nomenclature was to contrast the theory to those that suggest a more strict hierarchy with object-like representations only existing at certain levels of processing \citep{Hawkins2019ANeocortex}. While models developed from this theory have been shown to be capable of rapidly learning objects and performing recognition, this was limited to synthetic data-sets \citep{Lewis2019LocationsCells}. A similar concept was used to develop a model that was able to recognize objects from images, but this was limited to recall of memorized examples, rather than generalization to novel examples of an object class \citep{Bicanski2019ACells}.

\textbf{Machine Learning Approaches} Few shot learning is a large field, and prior work has addressed learning hand-written characters with various techniques \citep{Wong2015OnePatches, Lake2016, George2017ACAPTCHAs}, or demonstrated the benefits of memory-like mechanisms in the few-shot setting \citep{Santoro2016One-shotNetworks}. Our intent is not to present the current work as a strong solution to the problem of few-shot learning. However, the few-shot setting captures our biological motivation of humans learning rapidly from arbitrary feature sequences. As such, we use the few-shot experimental setting to evaluate the performance of our system. 

\section{Methods}

\subsection{Overview}

Our work builds on the sensorimotor system implemented in \citet{Lewis2019LocationsCells}, which in turn uses many of the algorithms employed in \citet{Hawkins2017}. In this paper we address two limitations of the network in \citet{Lewis2019LocationsCells}. First, they used synthetic objects and features rather than those derived from natural data-sets. Second, their system was designed to only recall previously seen objects and did not generalize to other examples of an object class. In order to remove those limitations, we make two main changes: 

\begin{itemize}
    \item We implement a sparse convolutional feature detector, trained on images. When the sensor moves to sample an image patch, the corresponding subset of sparse feature outputs is sent to the sensorimotor network.
    
    \item We enhance the classifier by storing multiple location representations from the training examples of each class. Classification then operates on unions of grid cell locations.

\end{itemize}

We describe each component in more detail below. Except where noted, the sensorimotor network is mathematically as described in \citet{Lewis2019LocationsCells}, and we advise readers interested in those details to refer to that work.

\subsection{Sparse Feature Extraction}\label{hyper_parameter_section}

\begin{figure}[ht]
\begin{center}
\includegraphics[width=0.9\textwidth]{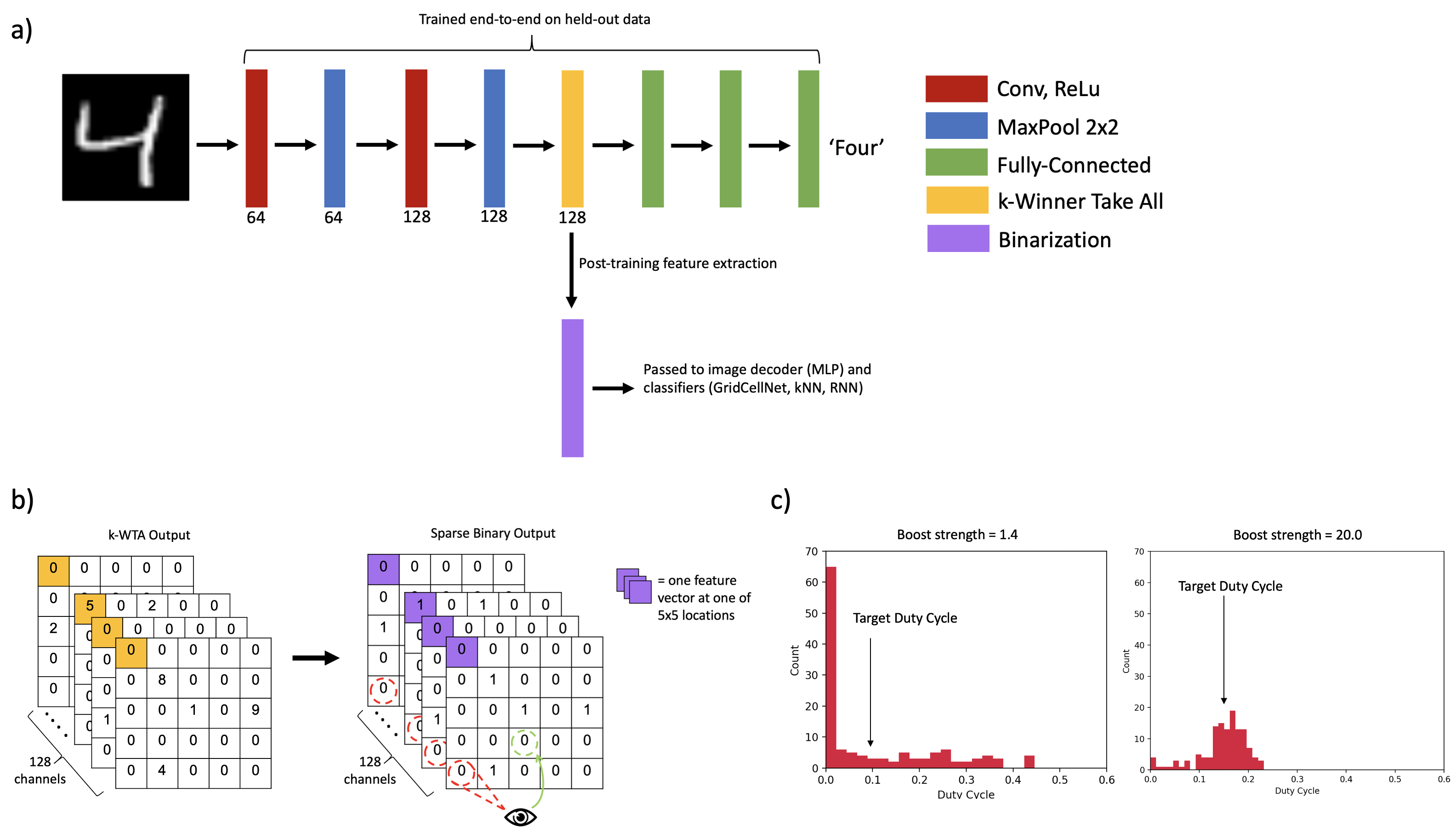}
\end{center}
\caption{\label{fig:cnn_architecture}\textit{The pre-processing convolutional neural network, including its use in generating the input features for later classifiers.} a) The encoder CNN is trained end-to-end to perform classification, with a k-WTA operation that constrains the mid-level representations to a specific level of sparsity. Numbers below operations show the channel dimension. b) An example of what the k-WTA and binarized representations might look like. Note that in all of our tasks, the classifiers are given a sparse feature vector (dimension 128) from each of the 5x5 locations in a sequence, represented with the eye and its movement. The order with which the features are sampled across this 5x5 space can either be fixed for all examples during both training and testing, or follow an arbitrary sequence. c) For optimal performance, the two parameters of k-WTA (target duty cycle and boost strength) are optimized so as to ensure most of the neurons achieve the target duty cycle. On the left is shown typical values for boosting factor used in past models, while on the right we show the result of using the larger boosting factor and target duty cycle that we arrived at through hyperparameter tuning.}
\end{figure}

In order to handle realistic images, we use a trained convolutional neural network to generate sparse binary features at multiple image locations. Specifically, we trained a convolutional neural network (CNN) \citep{Lecun1998LeNet} in a supervised paradigm on a subset of the MNIST training data-set of handwritten digits (54,000 images) \citep{LeCun1998Gradient-basedRecognition}, tuning it using a hold-out cross-validation section (6,000 images). This encoder network had the architecture shown in Figure \ref{fig:cnn_architecture}a, where the second max-pooling operation was followed by a k-Winner Take All layer (k-WTA) \citet {Ahmad2019HowRepresentations} to enforce sparsity in the representation (Figure \ref{fig:cnn_architecture}b). While the non-zero values in this layer took on real-number values (necessary for useful gradients during learning), we require binary feature vectors for input to the sensorimotor network. We therefore used the network after training to pass images through until the k-WTA layer, and then binarized this representation, providing us with a 5x5 grid of features. At each grid location we then have a vector of dimension 128 which contains regional information about the image representation in a sparse format. The feature vectors at each of these 25 locations form the input to all of our downstream classifiers. Note therefore that none of our classifiers (GridCellNet and the control comparisons) received direct pixel inputs as their features. 

\textbf{CNN Details} It is possible that binarization could lead to significant loss of information. We verified that a linear classifier can be trained to perform accurate classification with these as feature inputs with an accuracy of 99\%+. We also verified that a decoder could accurately reconstruct the input image. In order to achieve optimal performance, it was useful for the feature vector to have a reasonably large dimension (128),  number of non-zero elements (19), and high entropy (intuitively, how often did each input feature contribute to a representation across all examples). To optimize the entropy, we made use of the k-WTA's duty cycle (which monitors how often a unit is contributing to representations), and boosting factor (which biases a unit's activity to target a given duty cycle) (see \citet {Ahmad2019HowRepresentations} for details). This ensures a greater number of neurons each contribute information at some point, and the representation becomes more distributed.

In order of layers, the CNN architecture was composed of a convolution (kernel size 5, channels 64), max-pooling, convolution (kernel size 5, channels 128), max-pooling, k-WTA, and three fully connected layers (dimensions of 256, 128, and 10). k-WTA applied to the max-pooling layer was local (that is, the k-winners were determined across all channels at a given spatial location, rather than across the entire image space); this ensured that each extracted feature vector would have the same sparsity. We used stochastic gradient descent with a learning rate of 0.01, momentum 0.5, batch size of 128, and 10 epochs of training. 

\subsection{Sensorimotor Network}

The sensorimotor network consists of two layers, one representing the sensory input, the other representing the location of the sensor (Fig \ref{fig:general_intuition}c). Cells in both layers can be either on or off, and activity in the network proceeds through a series of discrete time steps. The sensory layer receives the input features, as well as modulatory input from the location layer. The location layer receives movement information, as well as input from the sensory layer. The connections between the sensory and location layers are modelled as dendritic segments, the small branches on biological neurons that integrate multiple synapses. A dendritic segment is deemed active if there is a significant match between the sending layer's sparse activity and sparse, learned weights. This match must exceed a user-set threshold, and these dendritic segments enable a given layer's activity to predict representations in the other layer. 

The basic intuition for inference is that at the outset, a sensory feature is likely to be ambiguous as to the nature of the object, and so the network should encode this ambiguity with a representation that corresponds to a union of all the objects compatible with this given feature. For example, a curved contour at the top of an image might represent a 9 or a 0 (Figure \ref{fig:recognition_algorithm}a). The object-representation is encoded by the activity in the location layer (as each learned object uses a unique location space), and so this union of multiple objects will correspond to multiple cells being active in each grid cell module. In our example, the active location representation will correspond to both where a curved contour was learned for a 9 and where it was learned for a 0. As additional features are sensed, the network will use its current representation of candidate objects to predict the next feature, with only those that are compatible with the subsequent sensation remaining part of the representation. Notably, these predictions rely on the presence of a feature \textit{at a given location}, and not simply a bag-of-features detector. If the sensor was to move to the bottom-left of the image, the learned 9 representation and 0 representation will predict different features. As more sensations are experienced, the system should converge to a specific representation that is consistent with a learned object. The object is recognised when the representation in the location layer corresponds to only that given object, and not any others. We now describe the network architecture and stages of inference in more detail. 

\textbf{Network Architecture} The location layer consists of 40 grid cell modules, each a lattice of 50 by 50 cells. A grid cell module has a particular scale and orientation, while the active location corresponds to the current phase of activity in the module. In our model, grid cells can be either active or inactive, and activation is determined by either the current representation in the sensory layer, or movement applied to the previous location representation; at time step $t$ and for grid cell module $i$, these are denoted by the binary arrays $A^{\tiny{\textrm{loc}},i}_{t,\tiny{\textrm{sense}}}$ and $A^{\tiny{\textrm{loc}},i}_{t,\tiny{\textrm{move}}}$ respectively. We model the location phase that determines $A^{\tiny{\textrm{loc}},i}_{t,\tiny{\textrm{move}}}$ using a square rather than the biologically motivated triangular lattice used in \citet{Lewis2019LocationsCells}, although this has no major consequence for the system.

The sensory layer is identical to that used in \citet{Hawkins2017}. The input features are binary vectors of length 128 with 19 active values (i.e. approximately 88\% sparsity). The sensory layer in turn consists of a corresponding 128 mini-columns, which receive the input features in a one-to-one fashion. Each mini-column in the sensory layer consists of multiple cells (here 32). This enables the mini-columns to use sparse activity to uniquely encode features associated with particular objects (i.e. location representations). The active cells in mini-column $i$ at time-step $t$ are denoted by the binary array $A^{\tiny{\textrm{in}}, i}_t$.

\textbf{Stage 1: Using Movement to Update the Location Representation} If the location layer has active cells, then each module uses the current movement information to compute a new set of active cells. Each module will apply a translation to its 50 by 50 activation pattern, according to the movement information. The translation vector is different in each module, and is determined by applying the following dilative rotation to the movement vector:

\begin{equation}
M_i = \frac{1}{s_i}\begin{bmatrix}
\cos(\theta_i) & -\sin(\theta_i) \\
\sin(\theta_i) & \cos(\theta_i)
\end{bmatrix}
\end{equation}

where $i$ denotes the particular grid cell module, $\theta$ its orientation, and $s$ its scale.

The translated 50 by 50 pattern will rarely align neatly with the original 50 by 50 cells, except in discrete environments. Typically each active cell in the pattern will land on the corner between four cells, so each active cell will activate up to four cells after the translation vector has been applied. During inference, this is indeed what happens, but during learning we allow the grid cell module to have more certainty about the current location. During learning, translating the active pattern will not increase the number of active cells; instead, the module's internal state includes a list of high-precision active phases, and the module applies the translation to those phases, rather than estimating those phases from the current set of active cells. This difference in the algorithm's behavior in inference and learning reflects the fact that binary representations will always lead to some spatial uncertainty during inference.

When inference begins for a new object, no location information is available - this will instead become available at stage 4, discussed below, and so inference proceeds to stage 2.

\textbf{Stage 2: Predicting Sensory Input with the Location Representation} The cells in the mini-columns have dendritic segments which receive activity from the location layer. If a dendritic segment is active (that is, a cell in the sensory layer is predicted by the activity of the location layer), then it is in a predictive state. Let the binary vector $\pi^{\tiny{\textrm{in}}}_t$ denote the sensory cells that have at least one active dendritic segment, and $\theta^{\tiny{\textrm{in}}}$ a dendritic threshold; then:

\begin{equation}\label{eq:dendrites_threshold}
    \pi^{\tiny{\textrm{in},c}}_t = \begin{cases}
    1, \exists_{d}[D^{\tiny{\textrm{in}}}_{c,d} \cdot A^{\tiny{\textrm{loc}}}_{t,\tiny{\textrm{move}}} \geq \theta^{\tiny{\textrm{in}}}] \\
    0, \textrm{otherwise}
    \end{cases}
\end{equation}

\textbf{Stage 3: Determining Activity in the Sensory Layer} If a given sensory layer cell that is predicted also receives activity from the input feature (that is, it is in a column receiving sensory input and was therefore correctly predicted), it will be active and inhibit any other cells in the mini-column that are not predicted. Note that multiple cells in any given mini-column can be active if they are predicted by the current location representation. If no cells in a mini-column are predicted but it receives sensory input, then all cells in the mini-column will become active. 

\textbf{Stage 4: Using the Sensory Representation to Update the Location} After the sensory representation has been determined, the location layer receives inputs from the sensory layer. In particular, the sensory features help to recall location information, and supplement the location representation arrived at by path integration. Similar to equation \ref{eq:dendrites_threshold}, this is determined by the overlap between the active cells in the sensory layer, and the learned weights. In this case however, an active dendritic segment is sufficient for a location cell to now be active, such that:

\begin{equation}\label{eq:location_dendrites_threshold}
    \pi^{\tiny{\textrm{loc},i,c}}_t = \begin{cases}
    1, \exists_{d}[D^{\tiny{\textrm{loc},i}}_{c,d} \cdot A^{\tiny{\textrm{in}}}_{t} \geq \theta^{\tiny{\textrm{loc}}}] \\
    0, \textrm{otherwise}
    \end{cases}
\end{equation}

\begin{equation}\label{eq:location_input}
    A^{\tiny{\textrm{loc}},i}_{t,\tiny{\textrm{sense}}} = \begin{cases}
    \pi^{\tiny{\textrm{loc},i}}_t, \left\Vert\pi^{\tiny{\textrm{loc},i}}_t\right\Vert > 0 \\
    A^{\tiny{\textrm{loc}},i}_{t,\tiny{\textrm{move}}}, \textrm{otherwise}
    \end{cases}
\end{equation}

At this stage, the next movement is received, and the four stages are repeated for the next time-step until inference is successful (discussed further below).

\textbf{Learning} Learning takes place via the reciprocal strengthening of connections between the active representation in the sensory layer, and the current location representation. The aim is to associate a given feature with a given location in that object's reference frame. When a new object is learned, the location representation at the first sensation is randomly initialized, such that each object operates in a different location space. This enables multiple objects to be jointly represented during inference, as the probability of overlap between different objects' location spaces is low. As further sensations are performed during learning, the location representation is updated using movement information as described in Stage 1 above. For the sensory layer, as the feature-location association has yet to be learned, a random cell in each mini-column receiving an input will be selected to be active. Each active cell in the sensory and location layer will then form reciprocal connections on one of their dendritic segments ($d'$) according to the following:

\begin{equation}
    D^{\tiny{\textrm{loc}}}_{c,d'} := D^{\tiny{\textrm{loc}}}_{c,d'}|A^{\tiny{\textrm{in}}}_{t,\tiny{\textrm{learn}}}
\end{equation}

\begin{equation}
    D^{\tiny{\textrm{in}}}_{c,d'} := D^{\tiny{\textrm{in}}}_{c,d'}|A^{\tiny{\textrm{loc}}}_{t,\tiny{\textrm{sense}}}
\end{equation}

Here "$|$" is used to indicate the bitwise OR operator; that is, if a synapse already exists between two cells, then it is unaffected by the learning rule. 

Following the above, the network can rapidly learn objects by visiting each feature once, performing a single set of weight updates for each feature. Note that due to the path-integration performed by the grid cells, both learning and inference can take place using an arbitrary order through the features of the object - there need be no correspondence between the order taken at learning vs. that used at testing. Note also that the learning process could in principle be implemented on hardware in a parallelized form, although for biological plausibility, this is run as a serial process. 

There are several hyperparameters of the model that might be tuned to optimize performance, such as increasing the grid cell module size to increase the capacity of the model (see \citet{Lewis2019LocationsCells} for a quantitative exploration of this). The main parameter we tuned was the dendritic threshold $\theta^{\tiny{\textrm{loc}}}$. If this was too high, grid cells were too stringent in which sensory features were present to become active; if it was too low, grid cells were too easily activated by spurious sensory features. Details on how we tuned hyper-parameters are provided further below.

\subsection{Classification}

\begin{figure}[ht]
\begin{center}
\includegraphics[width=1.0\textwidth]{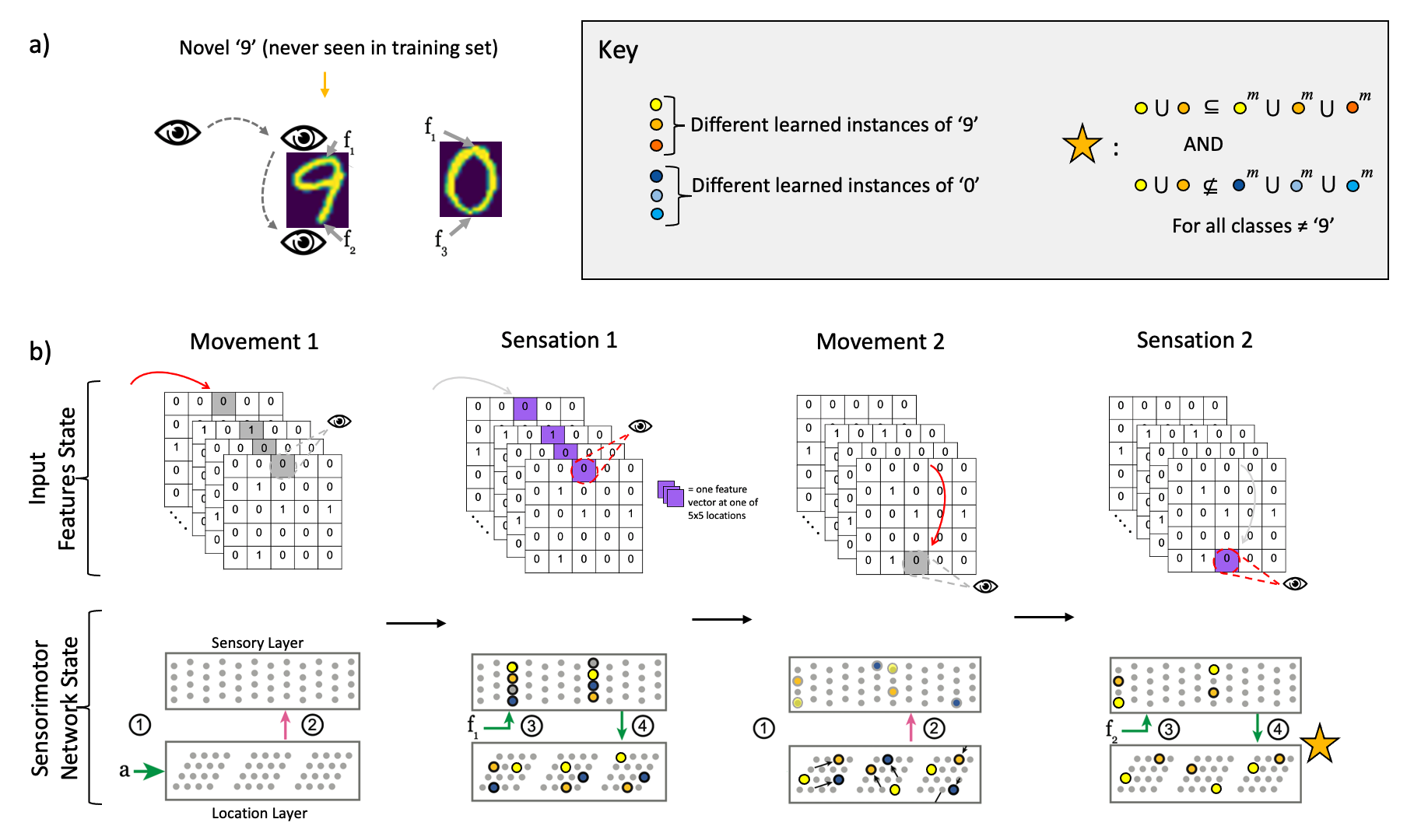}
\end{center}
\caption{\label{fig:recognition_algorithm}\textit{Overview of the GridCellNet classification algorithm.} a) The algorithm's concept at a high level. Note that a single sensation ($f_1$) is likely to be ambiguous as to the nature of the object, and so correct inference requires integration over several features. b) When the sensor moves to its first location (<1>, under Movement 1), there is no location representation on which to base sensory predictions (<2>). As a result, the first sensation (<3>), activates all the cells in columns that receive input. This activity then activates location representations which are consistent with that feature input (<4>). This will be a union of multiple object representations, some of which are consistent with the target class (shades of yellow), and some of which are not (shades of blue) - thus at this point the class identity of the object is ambiguous. With the next sensor movement (<1>, under Movement 2), the location representations of the grid cells are updated using path integration. The active grid cells then provide a prediction (via the modulatory impact of dendritic segments) to the sensory layer (<2>). The next sensory input (<3>) is consistent with two previously learned examples of 9's, i.e. with that feature representation at that location \textit{given the previous feature representations at the previous locations}. The remaining active location representations (<4>), are now a sub-set of the target class (the union of location representations of all learned 9's), and classification is successful. This step uses information about the relative position of the sensor, $m$, to constrain the comparison (e.g. the bottom-middle region of the unknown digit). The completion of inference is indicated with a star, although it may not be successful (discussed in main text). Note that Movement 1, Sensation 1, Movement 2, and Sensation 2 indicate the same two-layer network over the discrete time-steps of the algorithm. Figure adapted with permission from \citet{Lewis2019LocationsCells}. 
}
\end{figure}

In order to enable classification of unseen objects, we extended the classification algorithm in \citet{Lewis2019LocationsCells}. Our recognition algorithm is summarized in Figure \ref{fig:recognition_algorithm}. The original algorithm required the location representation of the object to be a subset of the target representation, where the target was a single learned example. In contrast, we designated inference as the location representation being a subset of the union of a particular class' possible location representations, but not any other's. That is, inference requires that the location representation is a subset of only one class. Additionally, the inference stage uses information about the current position of the sensor on the unknown object to constrain the location representations it compares to. Note that any location representation (either currently instantiated in the network, or previously learned), can be represented by the set of active grid cells. Formally, the classifier's probabilities for class $y$ during inference therefore corresponds to:

\begin{equation}
    p(y) = \begin{cases} 1,
    (A^{\tiny{\textrm{loc}}} \subseteq L_{i=y}^m) \textrm{ AND }
    (A^{\tiny{\textrm{loc}}} \not\subseteq L_{i\neq y}^m) \\
    0, \textrm{ otherwise}
    \end{cases}
\end{equation}

where $L_i^m$ is the joint location representation of all examples learned for class $i$ at the sensor position $m$. Here, the position information $m$ concerns the whereabouts of the sensor in the reference frame of the unknown object. Note that the network is able to develop a representation of the object and make predictions using the entirely internal reference frame provided by the grid cells. In order to perform the classification step however, learning and inference does require some notion of sensor position in the reference frame of the unknown object's shape ("Where on the digit that I'm trying to recognise is my sensor?"). Doing so ensures that the location representations which are compared to are sensibly constrained.

The above inference is successful if $p(y)=1$ and $y$ is the target class. Note however that there are three main failure cases. One is that when inference has taken place, $y$ is some other class that is not the target. In this case, the representation has converged to a wrong digit (for example converging to a 4 when it was in fact a 9). The second failure case is that the representation never converges to a subset of a learned representation (for example if a particularly unusual 6 is encountered). Finally, it is possible that the representation simultaneously satisfies multiple classes in the same inference step. This might be handled by random selection of one of the winning representations, but in our simulations this never occurred. 

In principle, a risk of this approach is that as the number of learned examples grows, the unions that form the target representations would become large and saturate the sparsity and capacity of the system; in practice we found the system was able to work well with few shot-learning (e.g. up to 100 examples per class), while we never explored training on the full data-set due to demanding wall-clock training times.

\subsection{Comparison Networks and Decoder Network}

We compare our architecture to both an RNN and a k-NN classifier. For our RNN, we used a long short-term memory (LSTM) classifier \citep{Hochreiter1997LongMemory}. This network received an input sequence of length 25, corresponding to the 25 locations in the image feature space. Each feature vector in this sequence was a sparse feature vector extracted from our CNN, as for the GridCellNet. In addition, the indexed position of this feature in the input sequence (in the form an integer from 0 to 24) was provided as an additional feature element (thus each input feature vector consisted of 129 elements). Whether the order with which the sequence was provided was fixed across objects was determined by the experimental condition, but the location information provided as the additional feature value always represented the ground truth index/location of the feature. 

The RNN had a single hidden layer of dimension 128. Using additional layers did not appear helpful. We used weight decay \citep{Ng2004FeatureInvariance} of 0.001 and optimized with Adam \citep{Kingma2015Adam:Optimization}. Learning rates were selected via a grid-search for the best performance on each classification task independently, where each task is distinguished by both the number of training examples, and whether evaluation was using a fixed or arbitrary input sequence. 

We also compared our network to a k-NN classifier. This received the same input as the RNN, but as a single extended array, rather than sequentially, and without the additional location information. The number of neighbours for the k-NN classifier \citep{Fix1989DiscriminatoryProperties} and the dendritic threshold $\theta^{\tiny{\textrm{loc}}}$ were similarly selected via a grid-search for each classification task. Additional details for how the data-set was divided and the selected hyper-parameters are provided in Appendix \ref{hyper_appendix}. 

In order to enable visualization of the current feature representations in the GridCellNet (including its predictions), we trained a multi-layer perceptron (decoder network) on the sparse binary features from a subset of the MNIST training data-set (54,000 images). This decoder had a single hidden layer of dimension 512, with input 128x5x5 and output 28x28. For the decoder we used Adam with a learning rate of 0.001, a batch size of 64, and 10 epochs of training.

\section{Experiments}
To assess the utility of our proposed network, we evaluate its accuracy in three main scenarios: a fixed input sequence, arbitrary input sequences, and partial input sequences. We also assess the ability of GridCellNet to make predictions of upcoming sensory inputs.

\subsection{Inference With Fixed and Arbitrary Sequences}

With a fixed input sequence, the sequence of samples across the space of 5x5 features follows the same order for all objects during both training and evaluation. An arbitrary input sequence more closely resembles active sensation and is far more challenging. In this scenario, the input order is randomly determined for every object, and is not fixed between training and testing. As the GridCellNet performs only a single weight update per feature of an object during learning, we compare to LSTMs with both 1 epoch of training as well as 50 epochs. Note therefore that our few-shot setting considers not only exposure to a limited number of training examples, but also limited opportunity for weight updates with each training example. 

\begin{figure}[ht]
\begin{center}
\includegraphics[width=0.9\textwidth]{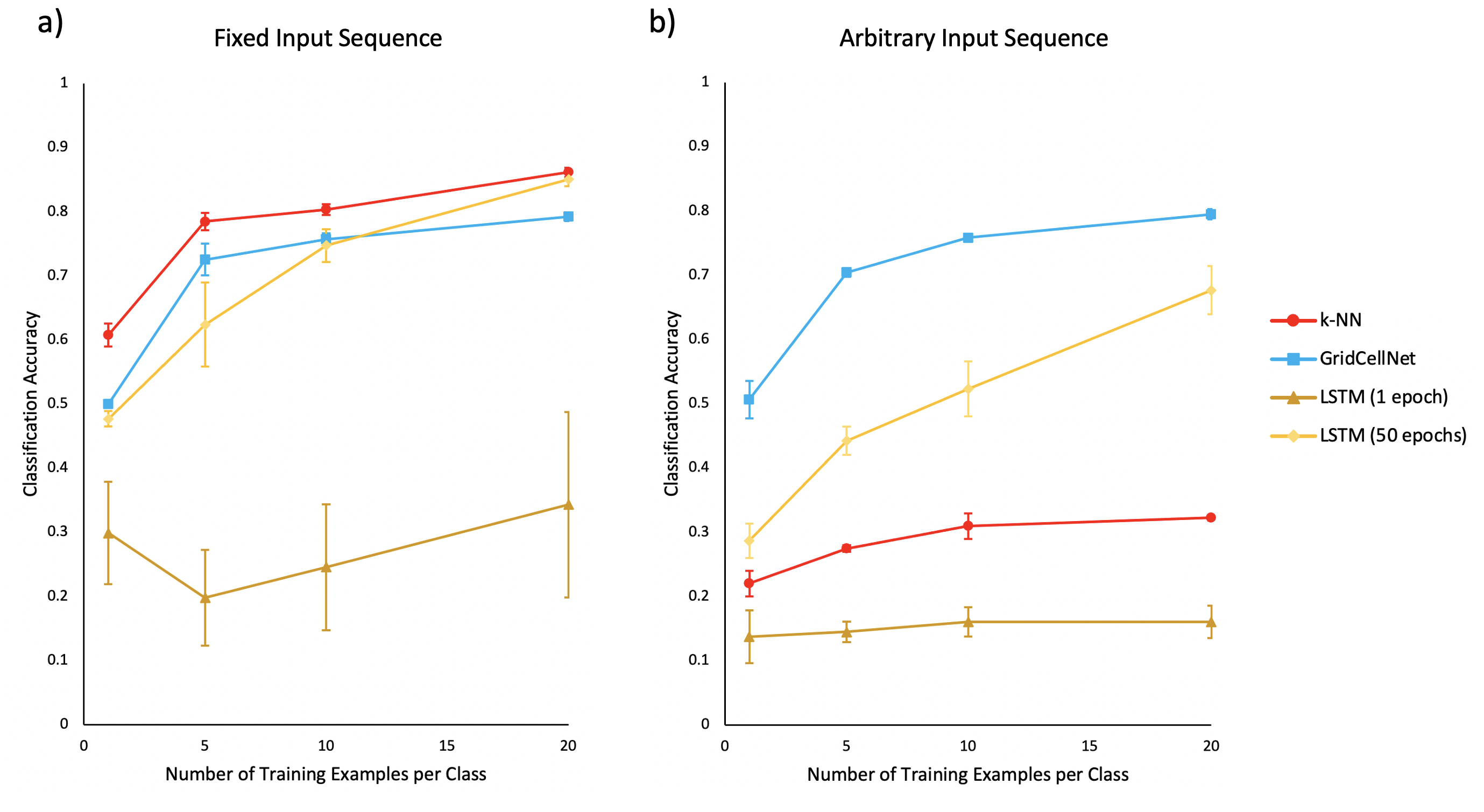}
\end{center}
\caption{\label{fig:main_accuracy_results}\textit{Performance of Classifiers Given Fixed or Arbitrary Sequences of Input.} Classification accuracy on 1000 examples of the MNIST test set as a function of the number of training examples per-class. a) Accuracy when an identical sequence of passes over the input space is used for both training and inference. b) Performance when the sequence can be arbitrary and different between training and inference. Error bars show the 95\% confidence interval of the mean across three random seeds.}
\end{figure}

We begin by evaluating the classifiers on the standard task of learning and generalization given a fixed input sequence. As expected, all of the classifiers do reasonably well (Figure \ref{fig:main_accuracy_results}a), with the exception of the LSTM constrained to only one weight update per image. GridCellNet's learning takes place using rapid Hebbian-like weight updates, and so unlike the LSTM, it can form robust representations in-spite of only having observed each training object once. 

We also note that the k-NN actually performs better than the LSTM with 50 epochs of training, despite the LSTM being provided with location information. This appears to be due to the challenge the LSTM faces of learning longer range dependencies given so few training examples, in spite of being provided with information about where the current feature is located in the sequence. The ability of the LSTM to solve this appears highly sensitive to adjustments in hyper-parameters including the learning rate. While additional hyper-parameter tuning might ameliorate this, it is notable that GridCellNet does not suffer from the issue of long-range dependencies, and performs at its optimum with minimal changes to its key hyper-parameter (see Table \ref{table:hyperparameters} in Appendix \ref{hyper_appendix}). 

We next assess performance in the setting of an arbitrary sequence input. As predicted, only GridCellNet maintains its performance (Figure \ref{fig:main_accuracy_results}b). In particular, the path integration properties of grid cells enable the network to represent the spatial location of features in a manner that can handle arbitrary and previously unseen movements through space. After perceiving a feature at a given location, the active grid cells perform path integration given the current movement to meaningfully update their representation of the new location. This in turn predicts the learned features at this new location. The particularities of the path that was taken is irrelevant to this process, and so the system is robust to arbitrary feature sequences.

It is worth noting that given sufficient training time and examples, the LSTM's performance steadily improves. Deep learning architectures are undeniably powerful (indeed we used them to perform the initial feature extraction step for all of our classifiers), and it is likely that given enough training examples, the LSTM's performance would match GridCellNet. Note in particular that GridCellNet uses only simple, Hebbian-like learning rules (i.e. no back-propagation of error). The contrast in performance in the few-shot setting however supports the proposed architecture as a principled and biologically plausible mechanism by which humans might rapidly learn under the challenging settings explored here. 

\subsection{Rapid Inference With Partial Input Sequences}

We have noted that, in principle, GridCellNet may successfully classify an object before it has received a complete sequence of all 25 features. Recall that classification has occurred as soon as GridCellNet's representation has converged to a subset of the representations associated with a particular class. 

To assess the performance of the GridCellNet as a function of the number of sensations, we simply determined the cumulative accuracy as a function of the number of sensations (total possible of 25). For k-NN, we iteratively fit k-NN classifiers with progressively more elements from the feature sequence available, and assessed their accuracy. 

In Figure \ref{fig:accuracy_vs_sensations}a, we show that GridCellNet classifies most of the examples given to it after observing only a fraction of the total input sequence; indeed, the majority of the successful classifications occur in under 10 sensations. For comparison we show that the k-NN with an arbitrary input sequence benefits little from additional examples - it is largely limited to key off of isolated features in order to perform at above chance levels. 

\subsection{Predictive Representations}

That GridCellNet can frequently perform inference after only a few iterations has an additional advantage beyond efficiency; due to the predictive nature of the network, it can represent features in unseen parts of the image.

To visualise the representation of the GridCellNet during inference, we extract the sequence of sensed and predicted features at each progressive sensation. As the sensor passes over the sequence of inputs, the networks representation accumulates the ground-truth sensations previously experienced, as well as the prediction for the next step. Importantly however, once inference is successful, all future accumulated representations are based solely on predictions from the inputs received up until the time of inference. The totality of these representations are then fed to the decoder network to visualise the output at different stages of inference. We show examples where the network converges to a single representation at inference time; the decoder in its current form is unable to make sense of the input if the representation at inference includes a union of object representations.

\begin{figure}[ht]
\begin{center}
\includegraphics[width=0.9\textwidth]{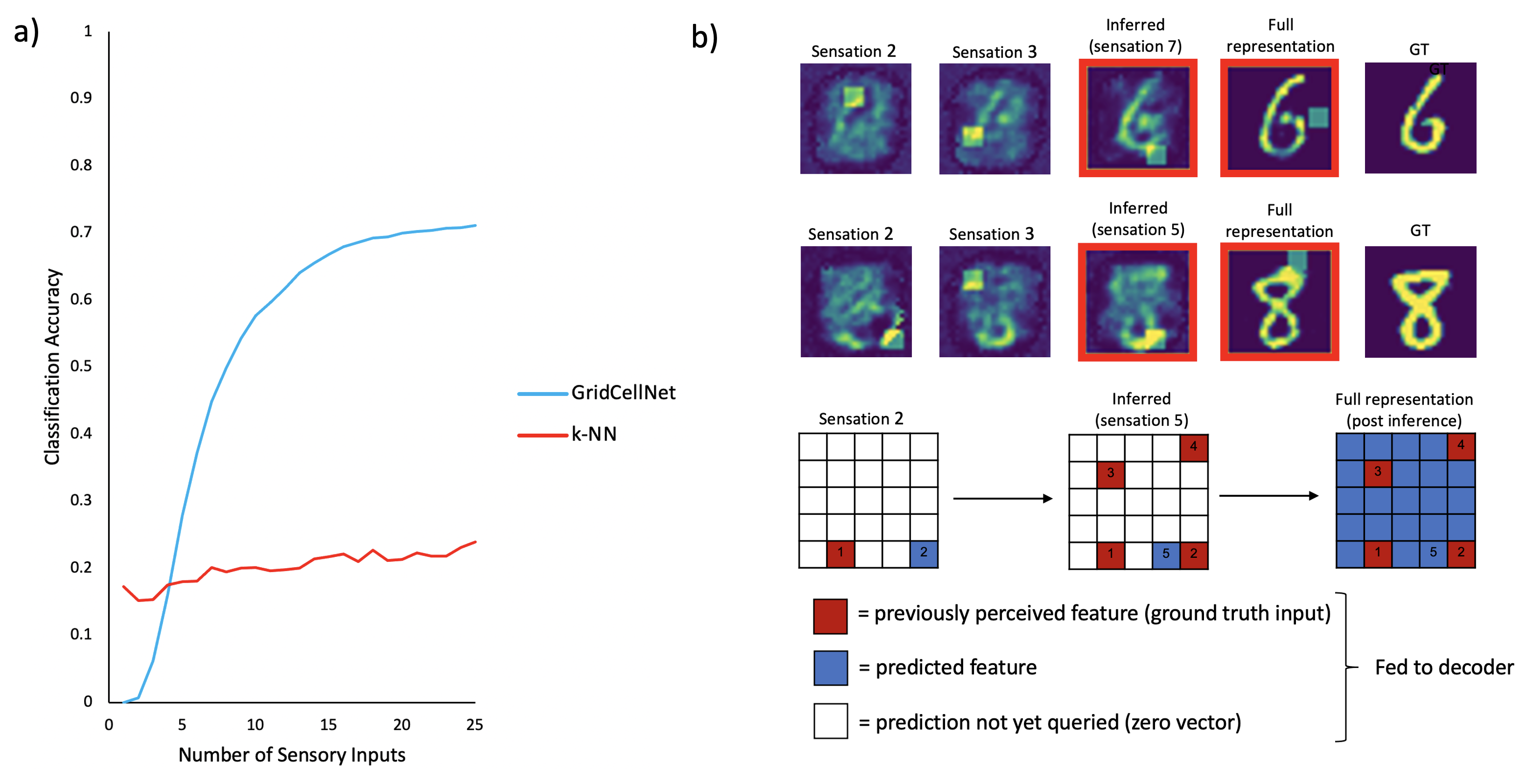}
\end{center}
\caption{\label{fig:accuracy_vs_sensations}\textit{Performance of GridCellNet as a Function of Sensations, and Predictive Capabilities.} Accuracy as a function of the number of sensations. Both classifiers were trained with 5 training examples per class, using arbitrary input sequences at training and test time. b) We use the sensed and predicted features of the GridCellNet and a decoder network to visualise the system’s representations. We show two example prediction sequences (top). GridCellNet predicts its next sensation (highlighted box) based on prior input. After successful inference (red border), the system predicts the entire image, substituting unseen features with those from memory. A diagram of the correspondence between past sensations and upcoming predictions is given at bottom, matching the representations seen for the `8' in the middle row. At each sequence step, the entire 5x5 grid of feature representations (each a vector of length 128) is fed to the previously trained decoder, and the corresponding reconstructed image shown. Note "Sensation 1" is not shown, as at this time, there is no prediction for the network to make.}
\end{figure}

Figure \ref{fig:accuracy_vs_sensations}b shows examples of GridCellNet restoring from memory an example that closely matches the input. In the early stages of inference, the internal representation consists of a small number of perceived features, and the next predicted feature (highlighted box). At inference time, enough features have been sensed that the decoded image often appears recognizable to a human. After then evaluating predictions at every unseen location, we observe that GridCellNet can recall an example from memory that is similar to the input.

Note that at every time-step, the model can make a prediction about any arbitrary location. It would therefore be possible in principle to query every location \textit{before} inference, but as the decoder we use has only been trained on single features (rather than a union of overlapping representations), these visualizations are not informative. 

\section{Discussion}

We have presented a novel approach to the challenge of active visual object recognition given an arbitrary sequence of feature inputs sampled across space. Robustness to this task is achieved through the use of grid cells to model the location of features in the reference frame of an object. In addition, our network takes advantage of rapid Hebbian-style weight updates to enable few-shot learning, and predictive components to enable the completion of partially observed images. 

This work was partly motivated by the observation that humans solve this task effortlessly when performing saccades, and that grid cells might enable a biologically plausible solution. It has been proposed that humans might perform object recognition in a variety of sensory modalities by making use of grid cell computations \citep{Hawkins2019ANeocortex}, including in vision \citep{Bicanski2019ACells}. In \citet{Bicanski2019ACells}, the authors used features extracted from multiple locations of an image to perform a task with some similarities to our own. These features were sequentially fed to a classifier that, similar to ours, integrated these features into a learned representation. When subsequently challenged to recall which of a handful of memorized images were presented, the system successfully did so, even under settings such as partial occlusion. Importantly however, their focus was on visual recognition \textit{memory}, and images used during training were the same as those used at evaluation time. Thus there was no need for the system to generalize to unseen examples, recognising the commonality between different instances of a class. Our work therefore represents the first demonstration that grid cell-like computations can be leveraged to enable generalization on a visual task to unseen examples of an object class. While this supports the plausibility of the human brain using such a mechanism, applying the proposed system to a more challenging data-set such as Omniglot \citep{Lake2016} is an important next step for demonstrating the capability of the approach.

In addition to the above relevance to neuroscience, the approach has implications for machine learning. In particular, providing a system with the flexibility to perform well with novel sampling sequences has an obvious advantage; assuming classification is possible without traversing the entire sequence space (as we demonstrated), then an agent using such a mechanism to perform object recognition could sample the most informative regions in a principled manner, and thereby operate more efficiently. The opposite to this is an agent that is constrained to sample every point, always following the same sequence such as a raster scan across the image. We emphasize that while the accuracy achieved by the GridCellNet is not as high as some other approaches to few-shot learning \citep{Wong2015OnePatches, Lake2016}, our intention is not to propose the model as a strong solution to that general setting. Rather our purpose is to show that in the context of few-shot learning, the use of grid cell representations can provide robustness to unpredictable inputs sequences, which might have downstream benefits for embodied agents. A future area of investigation will be pairing GridCellNet with a reinforcement learning agent that can learn to optimally control the movement of its sensor. It is also worth noting that the LSTM scales better with larger data-sets; as such, a sensible approach would likely be to combine the flexible, rapid learning of GridCellNet with longer-term learning in deep-learning architectures. 

Lastly, a natural benefit that the proposed system could bring is robustness to shifts of an object across an image (i.e. translation invariance). In spite of components intended to support translation invariance, CNNs still find this challenging \citep{Mu2019MNIST-C:Vision}. By relying primarily on representing features in an objects reference frame, we predict that the system would generalize to a novel location without any further training. We have not explored this however due to the constraints imposed by our feature extraction method, which does not guarantee the necessary equivariant shifts of features, as well as the current requirement that the classification criterion has some notion of where the sensor is in the reference frame of the unknown object. 

Despite the above promising results and avenues for future research, we must highlight several limitations of the current work. Although we attempted to ameliorate this somewhat by using separate data sub-sets, our method of feature extraction with a CNN is un-intuitive given our task of sequentially sampling feature patches. As noted above, it also constrains the application of GridCellNet to other interesting tasks such as image translation, and as such exploring alternative feature extraction methods such as a patch-based auto-encoder is an avenue for future research.

In order to support \textit{generalization} (rather than simply recall of a memorized example), there is currently a requirement that learning and classification receive information about where the sensor is positioned on the object being sensed. It is worth noting that this information was provided to the LSTM, and so it did not offer an unfair advantage for GridCellNet. Furthermore, all of the steps of developing an object representation (including prediction of unseen features), do not require this information, and the provided signal could be as simple as the sensor's position relative to the centre of mass of an object. The primary limitation with this requirement is that the network cannot make full use of its natural bias towards translation invariance. The requirement for positional information can be relaxed, in which case classification takes place by comparing the current location representation to any of those associated with a given class, but this can overwhelm the representational capacity of the current form of GridCellNet. Future modifications such as the use of multiple cortical columns may help to address this challenge.

While we compare our architecture to an LSTM, it is possible that transformer networks \citep{Vaswani2017AttentionNeed} would perform better in this setting. They currently represent the state of the art in many sequence based tasks, including visual tasks \citep{Dosovitskiy2020AnScale}, and explicitly encode positional information. It is worth noting however that, while not explored here, grid cells can in principle encode 3D structure \citep{Klukas2020EfficientCells}. Transformer networks already suffer from efficiency issues with long sequences, for which the introduction of a third dimension in the input representation would be problematic. The performance of transformer networks on our task and the generalization of GridCellNet to 3D objects will therefore be topics for future investigations. 

In summary, we believe that this work supports the notion that the brain may use grid cell computations when performing visual object recognition, and that this might underlie some of the visual tasks in which humans still outperform engineered systems. Future studies will aim to demonstrate this principle on more complex data-sets than the simple MNIST task explored here.

\section{Appendix}

\subsection{Hyper-parameter Selection} \label{hyper_appendix}

\begin{figure}[ht]
\begin{center}
\includegraphics[width=0.7\textwidth]{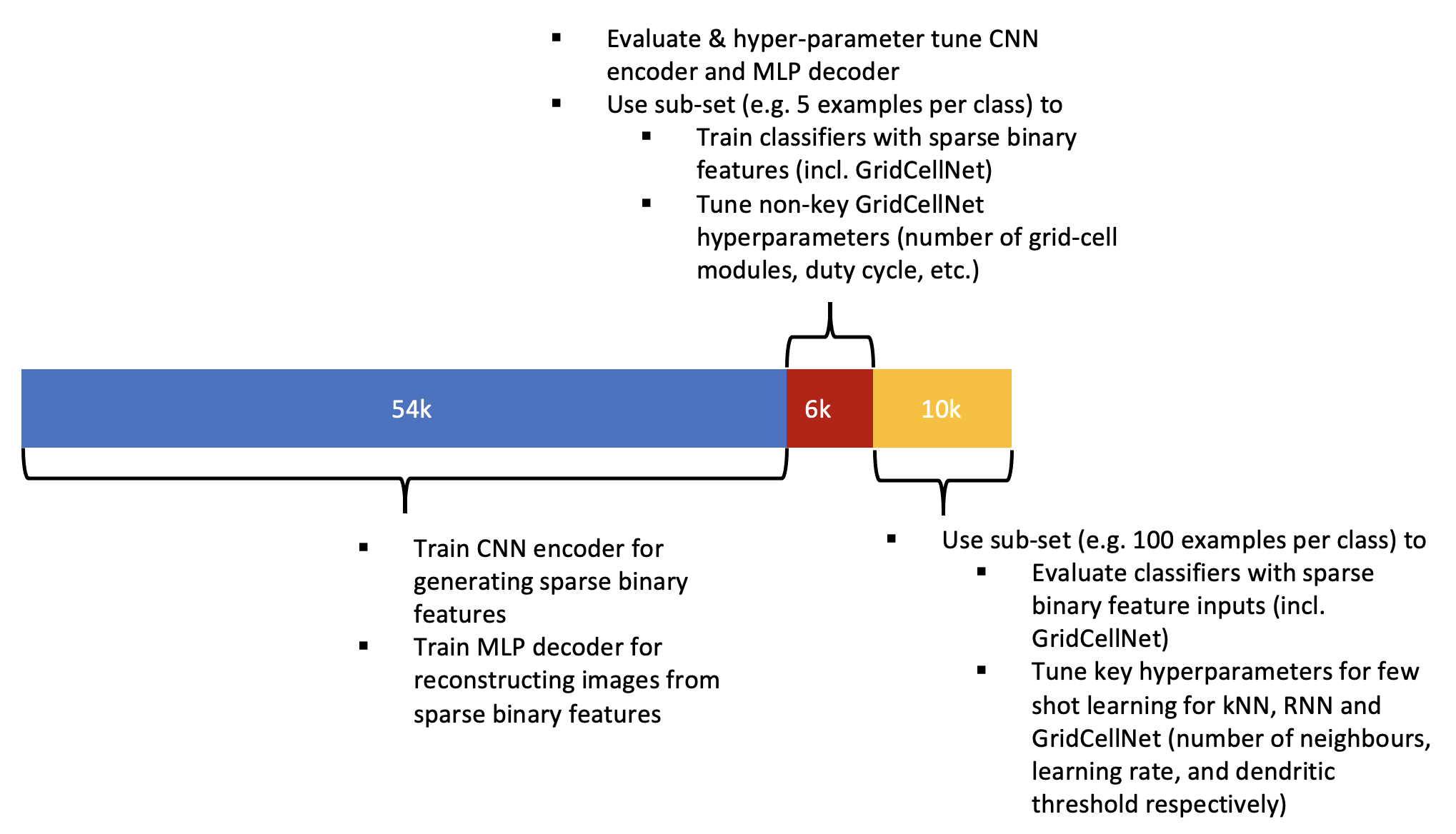}
\end{center}
\caption{\label{fig:data_splits}\textit{Division of the data-set for training and evaluation.} As the down-stream classifiers use features derived from a network classified end-to-end, and we are interested in few shot learning, we divide the data set such that the training of the various systems uses different sub-sets of the data. This also allows us to perform hyper-parameter tuning on hold-out data, with the exception of the key hyperparameters which are deliberately selected to enhance few-shot learning on the final evaluation data for all classifiers. Note that although images are therefore re-used during this step of hyper-parameter tuning and final evaluation, the feature vectors generated by the CNN encoder will vary with the random seed used, and therefore between the settings in which hyper-parameter tuning and evaluation are performed.}
\end{figure}

In Figure \ref{fig:data_splits}, we provide a break-down of how different splits of the data are used for different steps in feature extraction, hyper-parameter tuning, and learning. In Table \ref{table:hyperparameters}, we list the hyper-parameters arrived at for each classifier, specific to each learning task.

\begin{table}[t]
\caption{\label{table:hyperparameters}Choice of Hyper-parameters for Classifiers Given Fixed or Abitrary Input Sequences}
\label{sample-table}
\begin{center}
\begin{tabular}{l|cc}
     \# of training examples per class     & Fixed     &  Arbitrary  \\ \hline
          GridCellNet    &      Dendritic Threshold $\theta^{\tiny{\textrm{loc}}}$ &    \\ \hline
1          & 11                  & 11  \\
5          & 13                  & 13  \\
10          & 13                  & 13  \\
20          & 15                  & 15  \\ \hline
          LSTM (1 epoch)    &      Learning Rate    \\ \hline
1          & 0.005                  & 0.005  \\
5          & 0.005                  & 0.005  \\
10          & 0.005                  & 0.01  \\
20          & 0.01                  & 0.01  \\ \hline
          LSTM (50 epochs)    &      Learning Rate    \\ \hline
1          & 0.002                  & 0.002  \\
5          & 0.02                  & 0.002  \\
10          & 0.005                  & 0.005  \\
20          & 0.005                 & 0.01  \\ \hline
          k-NN    &      \# of Neighbours    \\ \hline
1          & 1                  & 1  \\
5          & 1                  & 1  \\
10          & 1                  & 7  \\
20          & 1                  & 9  \\
\end{tabular}
\end{center}
\end{table}


\begin{thebibliography}{31}
\providecommand{\natexlab}[1]{#1}
\providecommand{\url}[1]{\texttt{#1}}
\expandafter\ifx\csname urlstyle\endcsname\relax
  \providecommand{\doi}[1]{doi: #1}\else
  \providecommand{\doi}{doi: \begingroup \urlstyle{rm}\Url}\fi

\bibitem[Ahmad and Scheinkman(2019)]{Ahmad2019HowRepresentations}
Subutai Ahmad and Luiz Scheinkman.
\newblock {How Can We Be So Dense? The Robustness of Highly Sparse
  Representations}.
\newblock \emph{ICML 2019 Workshop on Uncertainty and Robustness in Deep
  Learning}, 2019.

\bibitem[Bicanski and Burgess(2019)]{Bicanski2019ACells}
Andrej Bicanski and Neil Burgess.
\newblock {A Computational Model of Visual Recognition Memory via Grid Cells}.
\newblock \emph{Current Biology}, 2019.
\newblock ISSN 09609822.
\newblock \doi{10.1016/j.cub.2019.01.077}.

\bibitem[Dosovitskiy et~al.(2020)Dosovitskiy, Beyer, Kolesnikov, Weissenborn,
  Zhai, Unterthiner, Dehghani, Minderer, Heigold, Gelly, Uszkoreit, and
  Houlsby]{Dosovitskiy2020AnScale}
Alexey Dosovitskiy, Lucas Beyer, Alexander Kolesnikov, Dirk Weissenborn,
  Xiaohua Zhai, Thomas Unterthiner, Mostafa Dehghani, Matthias Minderer, Georg
  Heigold, Sylvain Gelly, Jakob Uszkoreit, and Neil Houlsby.
\newblock {An Image is Worth 16x16 Words: Transformers for Image Recognition at
  Scale}.
\newblock \emph{arXiv preprint arXiv:2010.11929}, 2020.

\bibitem[Fiete et~al.(2008)Fiete, Burak, and Brookings]{Fiete2008WhatLocation}
Ila~R. Fiete, Yoram Burak, and Ted Brookings.
\newblock {What grid cells convey about rat location}.
\newblock \emph{Journal of Neuroscience}, 28\penalty0 (27), 2008.
\newblock ISSN 02706474.
\newblock \doi{10.1523/JNEUROSCI.5684-07.2008}.

\bibitem[Fix and Hodges(1989)]{Fix1989DiscriminatoryProperties}
Evelyn Fix and J.~L. Hodges.
\newblock {Discriminatory Analysis. Nonparametric Discrimination: Consistency
  Properties}.
\newblock \emph{International Statistical Review / Revue Internationale de
  Statistique}, 57\penalty0 (3), 1989.
\newblock ISSN 03067734.
\newblock \doi{10.2307/1403797}.

\bibitem[George et~al.(2017)George, Lehrach, Kansky, L{\'{a}}zaro-Gredilla,
  Laan, Marthi, Lou, Meng, Liu, Wang, Lavin, and Phoenix]{George2017ACAPTCHAs}
Dileep George, Wolfgang Lehrach, Ken Kansky, Miguel L{\'{a}}zaro-Gredilla,
  Christopher Laan, Bhaskara Marthi, Xinghua Lou, Zhaoshi Meng, Yi~Liu, Huayan
  Wang, Alex Lavin, and D.~Scott Phoenix.
\newblock {A generative vision model that trains with high data efficiency and
  breaks text-based CAPTCHAs}.
\newblock \emph{Science}, 358\penalty0 (6368), 2017.
\newblock ISSN 10959203.
\newblock \doi{10.1126/science.aag2612}.

\bibitem[Hafting et~al.(2005)Hafting, Fyhn, Molden, Moser, and
  Moser]{Hafting2005MicrostructureCortex}
Torkel Hafting, Marianne Fyhn, Sturla Molden, May~Britt Moser, and Edvard~I.
  Moser.
\newblock {Microstructure of a spatial map in the entorhinal cortex}.
\newblock \emph{Nature}, 436\penalty0 (7052), 2005.
\newblock ISSN 00280836.
\newblock \doi{10.1038/nature03721}.

\bibitem[Hawkins and Ahmad(2016)]{Hawkins2016WhyNeocortex}
Jeff Hawkins and Subutai Ahmad.
\newblock {Why Neurons Have Thousands of Synapses, a Theory of Sequence Memory
  in Neocortex}.
\newblock \emph{Frontiers in neural circuits}, 10, 2016.
\newblock ISSN 16625110.
\newblock \doi{10.3389/fncir.2016.00023}.

\bibitem[Hawkins et~al.(2017)Hawkins, Ahmad, and Cui]{Hawkins2017}
Jeff Hawkins, Subutai Ahmad, and Yuwei Cui.
\newblock {A Theory of How Columns in the Neocortex Enable Learning the
  Structure of the World}.
\newblock \emph{Frontiers in Neural Circuits}, 11\penalty0 (October):\penalty0
  1--18, 2017.
\newblock ISSN 1662-5110.
\newblock \doi{10.3389/fncir.2017.00081}.
\newblock URL
  \url{http://journal.frontiersin.org/article/10.3389/fncir.2017.00081/full}.

\bibitem[Hawkins et~al.(2019)Hawkins, Lewis, Klukas, Purdy, and
  Ahmad]{Hawkins2019ANeocortex}
Jeff Hawkins, Marcus Lewis, Mirko Klukas, Scott Purdy, and Subutai Ahmad.
\newblock {A framework for intelligence and cortical function based on grid
  cells in the neocortex}.
\newblock \emph{Frontiers in Neural Circuits}, 2019.
\newblock ISSN 16625110.
\newblock \doi{10.3389/fncir.2018.00121}.

\bibitem[Hochreiter and Schmidhuber(1997)]{Hochreiter1997LongMemory}
Sepp Hochreiter and Jürgen Schmidhuber.
\newblock {Long Short-Term Memory}.
\newblock \emph{Neural Computation}, 9\penalty0 (8), 1997.
\newblock ISSN 08997667.
\newblock \doi{10.1162/neco.1997.9.8.1735}.

\bibitem[Julian et~al.(2018)Julian, Keinath, Frazzetta, and
  Epstein]{Julian2018HumanGrid}
Joshua~B. Julian, Alexandra~T. Keinath, Giulia Frazzetta, and Russell~A.
  Epstein.
\newblock {Human entorhinal cortex represents visual space using a
  boundary-anchored grid}.
\newblock \emph{Nature Neuroscience}, 21\penalty0 (2), 2018.
\newblock ISSN 15461726.
\newblock \doi{10.1038/s41593-017-0049-1}.

\bibitem[Killian et~al.(2012)Killian, Jutras, and Buffalo]{Killian2012ACortex}
Nathaniel~J. Killian, Michael~J. Jutras, and Elizabeth~A. Buffalo.
\newblock {A map of visual space in the primate entorhinal cortex}.
\newblock \emph{Nature}, 491\penalty0 (7426), 2012.
\newblock ISSN 00280836.
\newblock \doi{10.1038/nature11587}.

\bibitem[Kingma and Ba(2015)]{Kingma2015Adam:Optimization}
Diederik~P. Kingma and Jimmy~Lei Ba.
\newblock {Adam: A method for stochastic optimization}.
\newblock In \emph{3rd International Conference on Learning Representations,
  ICLR 2015 - Conference Track Proceedings}, 2015.

\bibitem[Klukas et~al.(2020)Klukas, Lewis, and Fiete]{Klukas2020EfficientCells}
Mirko Klukas, Marcus Lewis, and Ila Fiete.
\newblock {Efficient and flexible representation of higher-dimensional
  cognitive variables with grid cells}.
\newblock \emph{PLoS Computational Biology}, 16\penalty0 (4), 2020.
\newblock ISSN 15537358.
\newblock \doi{10.1371/journal.pcbi.1007796}.

\bibitem[Lake et~al.(2016)Lake, Ullman, Tenenbaum, and Gershman]{Lake2016}
Brenden~M. Lake, Tomer~D. Ullman, Joshua~B. Tenenbaum, and Samuel~J. Gershman.
\newblock {Building Machines That Learn and Think Like People}.
\newblock \emph{Behavioral and Brain Sciences}, \penalty0 (2012):\penalty0
  1--101, 2016.
\newblock ISSN 14691825.
\newblock \doi{10.1017/S0140525X16001837}.

\bibitem[Lecun et~al.(1998)Lecun, Bottou, Bengio, and Ha]{Lecun1998LeNet}
Yann Lecun, Leon Bottou, Yoshua Bengio, and Patrick Ha.
\newblock {LeNet}.
\newblock In \emph{Proceedings of the IEEE}, number November, pages 1--46,
  1998.
\newblock ISBN 0018-9219.
\newblock \doi{10.1109/5.726791}.

\bibitem[LeCun et~al.(1998)LeCun, Bottou, Bengio, and
  Haffner]{LeCun1998Gradient-basedRecognition}
Yann LeCun, Léon Bottou, Yoshua Bengio, and Patrick Haffner.
\newblock {Gradient-based learning applied to document recognition}.
\newblock \emph{Proceedings of the IEEE}, 86\penalty0 (11):\penalty0
  2278--2323, 1998.
\newblock ISSN 00189219.
\newblock \doi{10.1109/5.726791}.

\bibitem[Lewis et~al.(2019)Lewis, Purdy, Ahmad, and
  Hawkins]{Lewis2019LocationsCells}
Marcus Lewis, Scott Purdy, Subutai Ahmad, and Jeff Hawkins.
\newblock {Locations in the neocortex: A theory of sensorimotor object
  recognition using cortical grid cells}.
\newblock \emph{Frontiers in Neural Circuits}, 2019.
\newblock ISSN 16625110.
\newblock \doi{10.3389/fncir.2019.00022}.

\bibitem[Long and Zhang(2021)]{Long2021AFormation}
Xiaoyang Long and Sheng~Jia Zhang.
\newblock {A novel somatosensory spatial navigation system outside the
  hippocampal formation}, 2021.
\newblock ISSN 17487838.

\bibitem[Long et~al.(2021)Long, Deng, Cai, Chen, and Zhang]{Long2021ACortex}
Xiaoyang Long, Bin Deng, Jing Cai, Zhe Chen, and Sheng-Jia Zhang.
\newblock {A compact spatial map in V2 visual cortex}.
\newblock \emph{bioRxiv}, 2021.

\bibitem[Moser et~al.(2008)Moser, Kropff, and Moser]{Moser2008PlaceSystem}
Edvard~I. Moser, Emilio Kropff, and May~Britt Moser.
\newblock {Place cells, grid cells, and the brain's spatial representation
  system}, 2008.
\newblock ISSN 0147006X.

\bibitem[Mu and Gilmer(2019)]{Mu2019MNIST-C:Vision}
Norman Mu and Justin Gilmer.
\newblock {MNIST-C: A Robustness Benchmark for Computer Vision}.
\newblock \emph{ICML 2019 Workshop on Uncertainty and Ro- bustness in Deep
  Learning}, 2019.

\bibitem[Nau et~al.(2018)Nau, Navarro~Schr{\"{o}}der, Bellmund, and
  Doeller]{Nau2018HexadirectionalCortex}
Matthias Nau, Tobias Navarro~Schr{\"{o}}der, Jacob~L.S. Bellmund, and
  Christian~F. Doeller.
\newblock {Hexadirectional coding of visual space in human entorhinal cortex}.
\newblock \emph{Nature Neuroscience}, 21\penalty0 (2), 2018.
\newblock ISSN 15461726.
\newblock \doi{10.1038/s41593-017-0050-8}.

\bibitem[Ng(2004)]{Ng2004FeatureInvariance}
Andrew~Y. Ng.
\newblock {Feature selection, L1 vs. L2 regularization, and rotational
  invariance}.
\newblock In \emph{Proceedings, Twenty-First International Conference on
  Machine Learning, ICML 2004}, 2004.
\newblock ISBN 1581138385.
\newblock \doi{10.1145/1015330.1015435}.

\bibitem[O'Keefe and Conway(1978)]{OKeefe1978HippocampalFire}
J.~O'Keefe and D.~H. Conway.
\newblock {Hippocampal place units in the freely moving rat: Why they fire
  where they fire}.
\newblock \emph{Experimental Brain Research}, 31\penalty0 (4), 1978.
\newblock ISSN 00144819.
\newblock \doi{10.1007/BF00239813}.

\bibitem[Santoro et~al.(2016)Santoro, Bartunov, Botvinick, Wierstra, and
  Lillicrap]{Santoro2016One-shotNetworks}
Adam Santoro, Sergey Bartunov, Matthew Botvinick, Daan Wierstra, and Timothy
  Lillicrap.
\newblock {One-shot Learning with Memory-Augmented Neural Networks}.
\newblock \emph{NeurIPS 2016 Deep Learning Symposium}, arXiv:1605.06065, 2016.

\bibitem[Stensola et~al.(2012)Stensola, Stensola, Solstad, Fr{\O}land, Moser,
  and Moser]{Stensola2012TheDiscretized}
Hanne Stensola, Tor Stensola, Trygve Solstad, Kristian Fr{\O}land, May~Britt
  Moser, and Edvard~I. Moser.
\newblock {The entorhinal grid map is discretized}, 2012.
\newblock ISSN 00280836.

\bibitem[Vaswani et~al.(2017)Vaswani, Shazeer, Parmar, Uszkoreit, Jones, Gomez,
  Kaiser, and Polosukhin]{Vaswani2017AttentionNeed}
Ashish Vaswani, Noam Shazeer, Niki Parmar, Jakob Uszkoreit, Llion Jones,
  Aidan~N. Gomez, Łukasz Kaiser, and Illia Polosukhin.
\newblock {Attention is all you need}.
\newblock In \emph{Advances in Neural Information Processing Systems}, volume
  2017-December, 2017.

\bibitem[Wong and Yuille(2015)]{Wong2015OnePatches}
Alex Wong and Alan Yuille.
\newblock {One shot learning via compositions of meaningful patches}.
\newblock \emph{Proceedings of the IEEE International Conference on Computer
  Vision (2015)}, 2015.

\bibitem[Yarbus(1967)]{Yarbus1967EyeObjects}
Alfred~L Yarbus.
\newblock {Eye Movements During Perception of Complex Objects}, 1967.

\end{thebibliography}
\end{document}